\documentclass[letterpaper, 10 pt, conference]{ieeeconf}
\IEEEoverridecommandlockouts
\title{\LARGE \bf Safe Continual Domain Adaptation after Sim2Real Transfer of Reinforcement Learning Policies in Robotics}

\author{Josip Josifovski$^{1}$, Shangding Gu$^{2}$, Mohammadhossein Malmir$^{1}$, Haoliang Huang$^{1}$, 
\\ Sayantan Auddy$^{3}$, Nicol\'as Navarro-Guerrero$^{4}$, Costas Spanos$^{2}$ and Alois Knoll$^{1}$
\thanks{$^{1}$ Technical University of Munich, Germany.}%
\thanks{$^{2}$ University of California, Berkeley, USA}%
\thanks{$^{3}$ Technische Universität Berlin, Germany}%
\thanks{$^{4}$ \href{https://www.l3s.de/}{L3S Research Center}, Leibniz Universit\"at Hannover, Germany.}%
\thanks{\noindent Josip Josifovski and Mohammadhossein Malmir have been financially supported by the A-IQ READY project, which has received funding within the Chips Joint Undertaking (Chips JU) -- the Public-Private Partnership for research, development, and innovation under Horizon Europe -- and National Authorities under grant agreement No.\ 101096658. 
}
}

\usepackage{cite}  
\usepackage{amsmath,amssymb,amsfonts}
\usepackage{mathtools}
\usepackage{algorithmic}
\usepackage{graphicx}
\usepackage{textcomp}
\usepackage{xcolor}
\usepackage{xfrac}
\usepackage{booktabs}
\usepackage{url}
\usepackage{hyperref}

\usepackage{macros} 
\usepackage{glossaries}
\usepackage{paralist}
\usepackage{balance} 

\usepackage{empheq}

\usepackage{amsmath}
\usepackage[caption=false, font=footnotesize]{subfig}

\usepackage{threeparttable}
\usepackage{diagbox}

\usepackage{epsfig,epsf,fancybox}
\usepackage{mathrsfs}
\usepackage{amssymb}
\usepackage{color}
\usepackage{multirow}
\usepackage{paralist}
\usepackage{verbatim}
\usepackage{galois}
\usepackage{boxedminipage}
\usepackage{accents}
\usepackage{stmaryrd}
\usepackage[bottom]{footmisc}
\definecolor{light-gray}{gray}{0.85}
\usepackage{colortbl}
\usepackage{makecell}
\usepackage{hhline}
\usepackage{mathtools}
\usepackage{xcolor}         
\usepackage{bbold}
\usepackage[utf8]{inputenc} 
\usepackage{url}            
\usepackage{booktabs}       
\usepackage{amsfonts}       
\usepackage{nicefrac}       
\usepackage{microtype}      
\usepackage{MnSymbol}

\newcommand{\g}{\mathbf g}

\newcommand{\ba}{\begin{array}}
\newcommand{\ea}{\end{array}}

\usepackage{cite}
\newacronym{ROS}{ROS}{Robot Operating System}

\hypersetup{
  pdftitle={Safe Continual Domain Adaptation after Sim2Real Transfer of
Reinforcement Learning Policies in Robotics},
  pdfauthor={Josifovski, Gu, Malmir, Huang, Auddy, Navarro-Guerrero, Spanos and Knoll},
  pdfsubject={Robotics (cs.RO); Artificial Intelligence (cs.AI)},
  pdfkeywords={sim2real, Reality Gap, Safe Reinforcement Learning, Continual Learning, Real-World Robotic Tasks},%
  colorlinks=true,
  linktoc=all,
  linkbordercolor=white,
  linkcolor=black,
  citecolor=black,
  urlcolor=black,
}

\begin{document}
\maketitle
\thispagestyle{empty}
\pagestyle{empty}
                                       
\begin{abstract}

Domain randomization has emerged as a fundamental technique in reinforcement learning (RL) to facilitate the transfer of policies from simulation to real-world robotic applications. Many existing domain randomization approaches have been proposed to improve robustness and sim2real transfer. These approaches rely on wide randomization ranges to compensate for the unknown actual system parameters, leading to robust but inefficient real-world policies. In addition, the policies pretrained in the domain-randomized simulation are fixed after deployment due to the inherent instability of the optimization processes based on RL and the necessity of sampling exploitative but potentially unsafe actions on the real system. This limits the adaptability of the deployed policy to the inevitably changing system parameters or environment dynamics over time. We leverage safe RL and continual learning under domain-randomized simulation to address these limitations and enable safe deployment-time policy adaptation in real-world robot control. The experiments show that our method enables the policy to adapt and fit to the current domain distribution and environment dynamics of the real system while minimizing safety risks and avoiding issues like catastrophic forgetting of the general policy found in randomized simulation during the pretraining phase. Videos and supplementary material are available at \href{https://safe-cda.github.io/}{https://safe-cda.github.io/}.

\end{abstract}


\section{Introduction}

 \begin{figure}[t]
    \centering
 \includegraphics[width=0.90\columnwidth]{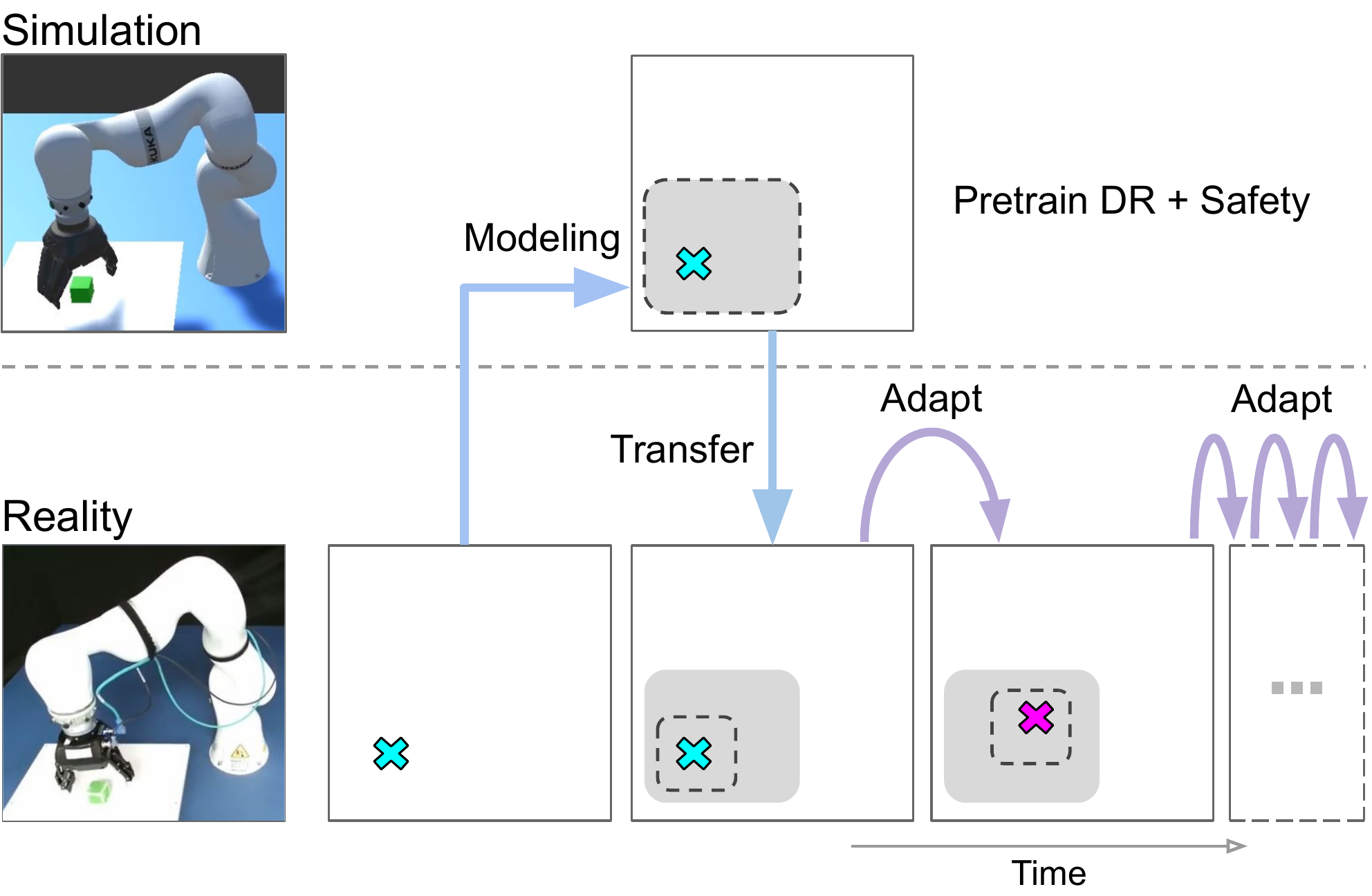}
	\caption{A conceptual representation of Safe Continual Domain Adaptation (SCDA). The unknown real system distribution (blue X) is approximated in simulation by randomizing the system parameters and training a robust RL policy (dashed square) covering wide parameter ranges, while simultaneously learning a safety Critic (shaded area). Then, the policy is adapted to the real system, using continual learning to prevent diverging from the initial robust policy and the safety Critic to avoid unsafe, exploitative actions in the real system. If real-world system parameters change (pink X) the policy can adapt to accommodate the change.}
    \label{fig:method}
    \vspace{-20px}
\end{figure}

Reinforcement Learning (RL) has demonstrated significant potential in finding high-reward policies for robotic systems. However, most results remain limited to simulation due to the inherent sample inefficiency of RL-based optimization and safety concerns. When real-world deployment is anticipated, RL policies are typically trained in highly parallelized simulators that either aim to precisely replicate the target domain, or introduce domain randomization to account for variability in system parameters \cite{muratore2022robot}. Both approaches seek to mitigate performance degradation caused by domain shifts between the simulator and the real system. Nevertheless, both strategies have inherent limitations. 

Specifically, developing exact simulations is often infeasible, as real-world system parameters may change over time, rendering previously measured parameters and the corresponding pre-trained policies obsolete. On the other hand, domain randomization can enhance policy robustness to a broad range of system variations, but it may also lead to suboptimal performance in simulation \cite{josifovski2022analysis} or result in an inefficient policy when deployed on the real system \cite{muratore2021data}. Even in cases where a sim2real transfer is successful, RL policies are typically deployed in a fixed manner, operating in inference mode. This is primarily due to challenges such as RL's instability during online optimization, sample inefficiency, biased data collection under the current policy, and safety concerns associated with exploitative yet potentially unsafe actions in real-world environments. However, deploying a fixed policy contradicts the reality that system parameters inevitably change over time. For instance, periodic variations in temperature or lighting may occur daily or seasonally, while gradual degradation due to wear and tear can affect long-term performance. Additionally, uncorrelated and unpredictable factors may introduce temporary or permanent deviations in system behavior. Thus, to achieve truly autonomous and adaptive robotic systems, developing policies that can continuously adapt after deployment is imperative, ensuring resilience to evolving real-world conditions.

Here, we propose a Safe Continual Domain Adaptation (SCDA) framework that leverages domain randomization with safe RL and continual learning (CL) to achieve robust policies in simulation while enabling continual adaptation to real-world systems. Our approach aims to mitigate safety risks and address challenges such as overfitting and catastrophic degradation or forgetting when adapting the policy after deployment. The overall methodology is illustrated in Fig.~\ref{fig:method}. We conduct experiments on policy transfer and adaptation for two robotic tasks, evaluating performance in realistic simulations and on physical robotic systems. Furthermore, we systematically analyze the impact of different components of our framework on policy adaptation to the target domain and dynamics, identifying key factors and trade-offs that influence real-world deployment. Our results demonstrate the benefits of adaptive policies and emphasize the potential of safe and continual RL adaptation methods in advancing autonomous robotic systems.

\section{Background and Related Work}

The reality gap problem and sim2real (sim-to-real) transfer \cite{jakobi1995noise} have remained active research areas. With advancements in deep learning for computer vision, early research based on supervised learning \cite{Tobin2017Domain, josifovski2018object} demonstrated that synthetic data could effectively be utilized to train models transferable to real-world data domains. At the same time, improvements in physics simulation and computational speed have facilitated the application of RL for robotic policy training and sim2real transfer \cite{Zhao2020SimtoReal}.
Though some studies have focused on developing highly precise simulations to ensure successful policy transfer \cite{kaspar2020sim2real}, randomized simulations have been more widely adopted, as this approach provides a more generalizable solution by eliminating the need for precise system identification and enabling the training of more robust policies in simulation \cite{muratore2022robot}. However, strong domain randomization may lead to suboptimal policy performance\cite{josifovski2022analysis}, highlighting the need for improved randomization strategies. Several techniques have been proposed to enhance domain randomization, including zero-zero shot transfer strategies like active domain randomization \cite{Mehta2020active}, adaptive randomization \cite{OpenAI2019Solving}, or strategies that use real-world data \cite{tiboni2023dropo}.

Moreover, ensuring safety is a critical challenge in deploying or adapting RL policies to real-world robotic systems. Reward shaping is a straightforward approach to incorporating safety in RL, where safety constraints are embedded in the reward function. However, this method introduces complexities in reward design, lacks a principled mechanism for enforcing safety constraints \cite{gu2024safe}, and does not provide formal safety guarantees. Instead, a more rigorous approach involves incorporating a cost Critic during training, which evaluates the safety of visited states and helps policies mitigate the risk of unsafe or potentially damaging situations. Various methods have been developed to address safety in RL \cite{gu2024review}, including single-agent constrained RL techniques such as constrained policy optimization \cite{achiam2017constrained} and reward-constrained policy optimization \cite{tessler2018reward}. In the multi-agent setting, multi-agent constrained policy optimization and multi-agent proximal policy optimization with a Lagrangian framework have been explored \cite{gu2023safe}. However, despite their theoretical and empirical validation in simulation, most of these methods lack extensive real-world experimentation, which may limit their practical deployment in safety-critical robotic applications.

Most sim2real approaches do not consider safety directly and only focus on policy adaptation during initial training but not after transfer \cite{valassakis2020crossing, tiboni2023dropo, huang2023what, muratore2022neural}.
In contrast, CL approaches \cite{parisi2019continual, wang2024comprehensive} aim to continuously adapt to evolving data distributions and environmental conditions over time. However, most CL methods assume that domain shifts occur in discrete steps, with well-defined task boundaries. CL in continuous data streams, where changes occur gradually, remains an active research area \cite{aljundi2019task}.
While CL techniques have been predominantly applied to computer vision problems \cite{parisi2019continual, wang2024comprehensive}, there is growing interest in their potential for RL \cite{khetarpal2022towards, schopf_hypernetwork-ppo_2022, abel2023definition} and robotics \cite{lesort2020continual, auddy2023continual, auddy2023scalable}. 

While most of the aforementioned works focus on a single aspect of the problem—such as sim-to-real transfer, domain randomization, domain adaptation, safe RL or CL—some related studies have explored combinations of these components. For example, DARAIL \cite{guo2024off} integrates domain shifts, RL, and adaptation by employing imitation learning and reward augmentation to improve model transfer from the source to the target domain. However, this approach does not account for safety or real-world adaptation over time. Similarly, Josifovski, Auddy et al.\ \cite{josifovski_auddy2024continual} combine CL, RL, and domain randomization during training to decouple randomization parameters and facilitate training under strong randomization. Nevertheless, this work does not consider safety aspects or adaptation in real-world deployment.
Other studies have focused on RL adaptation in real-world settings. For instance, Yin et al.\ \cite{yin2025rapidly} combine RL with real-world fine-tuning to improve policy performance after pretraining in simulation. Nonetheless, this method does not address safety concerns or the challenge of preserving the generality of the pre-trained model. Ganie et al.\ \cite{ganie2024online} integrate CL, safe RL, and adaptation for mobile robot formations, where safety is enforced through barrier functions rather than being learned. However, this approach does not explicitly tackle sim-to-real transfer. 

Furthermore, safe RL and adaptability have also been explored in non-stationary environments within a meta-learning framework \cite{chen2021context} where the focus is on finding robust and safe policies but is limited to simulation experiments. Cao et al.\ \cite{cao2024simplex} combine safe RL and CL using a coordination mechanism between an RL student and a model-based safety-verifying teacher to enable continual adaptation and sim-to-real transfer. This approach relies on a verifiable model-based component requiring domain knowledge.

In contrast, our proposed method integrates CL with safe RL. It enables robot learning safety in simulation and real-world control while mitigating catastrophic performance collapse over time. It uses a model-free safe policy search for robot learning in a randomized simulation at the start and real-system adaptation over time under domain shifts. 

\section{Methodology}
\subsection{Problem Formulation}\label{sec:problem}
Let $\Phi$ denote a real system with unknown dynamics, and $\mathcal{E} = \{\epsilon_1, \cdots, \epsilon_n\}$ denote the set of all randomization parameters of a simulator $\Sigma$ of $\Phi$. 

An RL agent optimizes a parameterized policy $\pi_\rho(a_t \mid s_t)$ under a Constrained Markov Decision Process (CMDP), defined as $(\mathcal{S}, \mathcal{A}, P, r, \mathbf{c}, \mathbf{b}, \gamma)$, where $\mathcal{S}$ denotes the state space, $\mathcal{A}$ denotes the action space, $P: \mathcal{S} \times \mathcal{A} \times \mathcal{S} \to [0, 1]$ denotes the transition probability, and $r: \mathcal{S} \times \mathcal{A} \to \mathbb{R}$ represents the reward function. In addition to the reward function $r$, we define a cost function $\mathbf{c} = (c_1, \dots, c_n) : \mathcal{S} \times \mathcal{A} \to \mathbb{R}^n$ and a threshold $\mathbf{b} = (b_1, \dots, b_n) \in \mathbb{R}^n$.
Within the simulation environment $\Sigma$, the agent aims to maximize the expected (discounted) cumulative reward for a given initial distribution $\rho$ while satisfying constraints on the expected (discounted) cumulative cost, i.e.,
\begin{equation}\label{eq:cmdp}
\begin{aligned}
\max_{\pi\in \Pi}\  V^\pi_r(\rho), \quad \text{s.t.} \quad V^\pi_{c_i}(\rho) \leq b_i, \quad \forall i =1,\ldots,n,
\end{aligned}
\end{equation}

where the expectation is taken over all possible trajectories, and $V^\pi_r(\rho)$ and $V^\pi_{c_i}(\rho)$ denote the value functions corresponding to the reward and cost functions, respectively. Similarly, we have the state-action function $Q(s, a)$ and the advantage function $A(s, a)$ for reward and cost \cite{gu2024balance}. 

Starting from the policy parameters $\tilde{\rho}$ optimized in a randomized simulation $\Sigma$, the policy is further adapted  to the real system $\Phi$. Here, the agent aims at finding a parameterized policy $\pi_\rho*(a_t|s_t)$ that maximizes the expected (discounted) cumulative reward on the real system. 

\subsection{Safe Continual Domain Adaptation}\label{sec:method}
To implement SCDA, we leverage the PCRPO algorithm \cite{gu2024balance} with Elastic Weight Consolidation (EWC)~\cite{kirkpatrick_overcoming_2017} Here, PCRPO is first used to learn safe policy under randomized simulation, and the final policy learned in simulation is further adapted on the real system with CL using EWC.

\subsubsection{Pretraining stage}
During the pretraining stage in simulation, we use wide randomization ranges, which expose the agent to higher uncertainty and unsafe states, in order to learn a robust and safe policy on the wide domain.  

PCRPO \cite{gu2024balance} divides policy optimization into three stages: a safety violation stage, a soft constraint violation stage, and a no-violation stage. A soft constraint violation region is defined around the safe constraint $b$, represented by $h^-, h^+$, forming the range: 
$$
[h^- + b, h^+ + b].
$$

A safety violation stage occurs when the cost value exceeds $ h^+ + b,$ while a no-violation stage corresponds to cost values below: $
h^- + b. $
Specifically, based on PCRPO \cite{gu2024balance}, we update the policy $\pi$ using Equation (\ref{eq:update-policy-parameters-for-safety}) with the cost gradient $\g_c = -\nabla V^\pi_c(\rho)$ and learning step size $\eta$ during a safety violation stage, where the policy $\pi$ is parameterized by neural network parameters $\tilde{\rho}$.
\begin{align} 
\label{eq:update-policy-parameters-for-safety}
\tilde{\rho}_{t+1}=\tilde{\rho}_t + \eta \g_c.
\end{align}

In a no-violation stage, where only reward optimization is required, we apply Equation (\ref{eq:update-policy-parameters-for-reward}) to optimize the policy using the reward gradient $\g_r = \nabla V^\pi_r(\rho)$.
\begin{align} 
\label{eq:update-policy-parameters-for-reward}
\tilde{\rho}_{t+1}=\tilde{\rho}_t+\eta \g_r.
\end{align}

In a soft violation stage, there are two cases. When the angle $\theta_{r,c}$ between the reward gradient $\g_r$ and the cost gradient $\g_c$ exceeds $90^\circ$, we define this as  a gradient conflict; otherwise, no gradient conflict occurs. In the presence of a gradient conflict, we update the policy using Equation (\ref{eq:projection-gradient-one}), where $x_t^r$ and $x_t^c$ are positive gradient weight parameters satisfying $x_t^r + x_t^c = 1$ for all $t \in T$. We update the policy using Equation (\ref{eq:projection-gradient-two}) if no gradient conflict exists.
\begin{align}
\label{eq:projection-gradient-one}
    \tilde{\rho}_{t+1}= \tilde{\rho}_t 
 + \eta \left[x_t^r \left(\g_r - \frac{\g_r \cdot \g_{c}}{\|\g_{c}\|^2} \g_{c}\right) + x_t^c \left(\g_{c} - \frac{\g_c \cdot \g_{r}}{\|\g_{r}\|^2} \g_{r}\right) \right], 
\end{align}
\begin{align}
\label{eq:projection-gradient-two}
    \tilde{\rho}_{t+1}= \tilde{\rho}_t + \eta \left[x_t^r \g_{r}+x_t^c\g_{c} \right].
\end{align}

Based on the gradient, we can have the loss function $\mathcal{L}_{\text{PCRPO}}(\rho) = \mathbb{E}_{s,a \sim \pi_{\tilde{\rho}{\text{old}}}} \left[ \frac{\pi_{\tilde{\rho}}(a|s)}{\pi_{\tilde{\rho}{\text{old}}}(a|s)} A(s, a) \right]$, where $A(s, a) = Q(s, a) - V(s)$ is the advantage function, and the advantage function can be computed via different reward and cost update stages \cite{gu2024balance}.

\subsubsection{Adaptation Stage}
After completing the pretraining stage, in the randomized simulation, we evaluate the parameterized policy $\tilde{\rho}$ to determine parameter importance based on EWC \cite{kirkpatrick_overcoming_2017}. Namely, we use the Fisher Information Matrix (FIM) to estimate the posterior distribution over the network parameters $\tilde\rho$, approximating it as a multivariate Gaussian centered around the maximum a posteriori (MAP) estimate. To calculate FIM, we roll out trajectories of the PCRPO agent in the randomized simulation while saving them in a replay buffer, and determine the parameter importance based on the (diagonal) empirical Fisher information using the squared gradients of the log-likelihoods of the action~\cite{kessler2020unclear} predicted by the PCRPO policy: 
\begin{equation}
    F(\tilde\rho) = \frac{1}{n} \sum_{i=1}^n \nabla_{\tilde\rho} \log p_{\tilde\rho}(y_i \mid x_i) \cdot \nabla_{\tilde\rho} \log p_{\tilde\rho}(y_i \mid x_i)^\mathrm{T}
    \label{eq:ewc_fisher}
\end{equation}
\noindent where $n$ denotes the number of samples in the replay buffer, $x_i$ and $y_i$, denote the input and output of the policy network, and $\tilde\rho$ denotes the parameters of the policy network.

After determining parameter importance, we continue training the agent on the real system. The loss function for adaptation on the real system after learning the task in simulation is:
\begin{equation}
    \mathcal{L}(\rho) = \mathcal{L}_\text{PCPRO}(\rho) + 
     \cfrac{\lambda}{2}  ||\rho -\tilde\rho||^2_{F\tilde\rho}
     \label{eq:ewc_loss}
\end{equation}
where $\mathcal{L}_\text{PCPRO}$ is the PCPRO loss as described earlier, $\lambda$ is a regularization constant, $\tilde\rho$ represents a snapshot of the policy parameters after pretraining, and ${F}_{\tilde\rho}$ is the diagonal Fisher information matrix for the policy parameters in the randomized simulation. This enables continual policy adaptation to the real system's subdomain and dynamics while preventing changes in parameters deemed important under the wide domain. The value and cost networks are not directly regularized and continue adapting better to represent the real system's reward and cost distribution. However, they are indirectly regularized through the sampling actions from the regularized actor.
\section{Experiments}
There are several questions that we want to answer through empirical evaluation:

\begin{itemize}
  \item Q1: Can SCDA improve the performance in the transferred domain w.r.t.\ the model pre-trained in randomized simulation and deployed in zero-shot transfer?
  \item Q2: Can SCDA handle sequential adaptation on a sub-distribution of the domain without forgetting the general policy?
  \item Q3: How do different components of the method (safety or reward-related adaptation, CL regularization) affect the performance?
\end{itemize}

\begin{figure}[t]
    \centering
	\includegraphics[width=0.80\columnwidth]{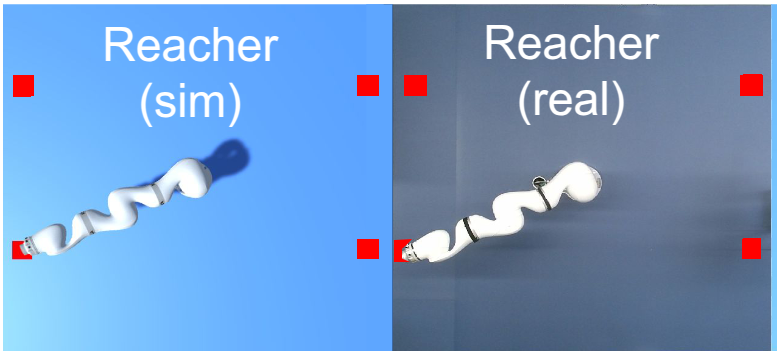}
	\caption{Reaching task -- simulation and real system environment.}
    \label{fig:reacher_env}
\end{figure}

To address these questions, we use a reach-and-balance task \cite{josifovski2022analysis}, where the RL agent has a closed-loop continuous control over the robot's joint velocities, and it is required to learn to balance between speed and safety to perform the task correctly. Formally, the state of the agent each timestep is defined as:
\begin{equation}
\mathbf{s}_t = [\Delta{x}_t, \Delta{y}_t, \Delta{z}_t,{q}_1,{q}_2] \in \mathbb{R}^5
\end{equation}
\noindent where $\Delta{x}$, $\Delta{y}$ and $\Delta{z}$ is the signed distance between the end-effector and the current target along each of the coordinate axes, while ${q_1}$ and ${q_2}$ are the joint positions. The actions at each timestep are 
\begin{equation}
\mathbf{a}_t = [\dot {q}_1,\dot{q}_2] \in \mathbb{R}^2
\end{equation}
\noindent where $\dot {q_i}$ are the requested joint velocities.

We independently optimize reward and safety-related cost functions. Namely:
\small
\begin{align}
\mathbf{r}_t &=
    \begin{cases}
    - \mathbf{d}_t   & \text{if } \neg S \\
    0 & \text{if } S
    \end{cases}
    \in \mathbb{R}, \quad
\mathbf{c}_t =
    \begin{cases}
    0   & \text{if } \neg S \\
    \mathbf{d}_t \cdot (\mathbf{T} - \mathbf{t}) & \text{if } S
    \end{cases}
    \in \mathbb{R}
\end{align}
\normalsize

\begin{figure*}[t]
    \centering
    \subfloat{\includegraphics[width=\textwidth]{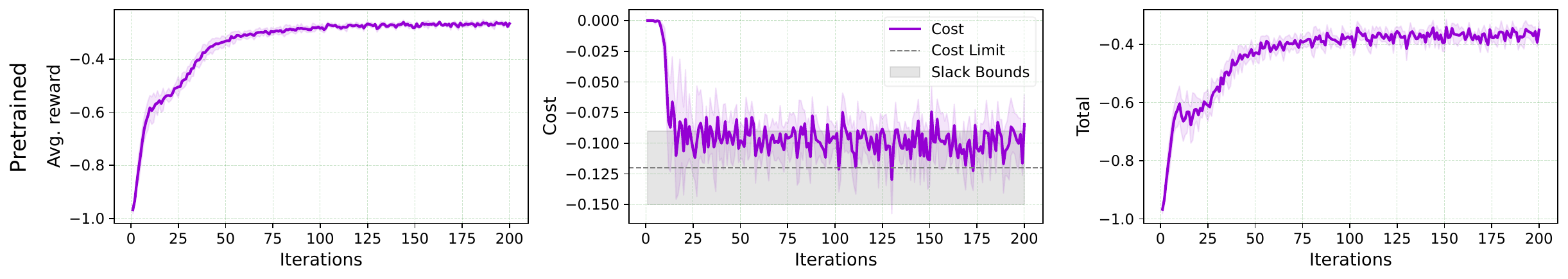}}\\
    \subfloat{\includegraphics[width=\textwidth]{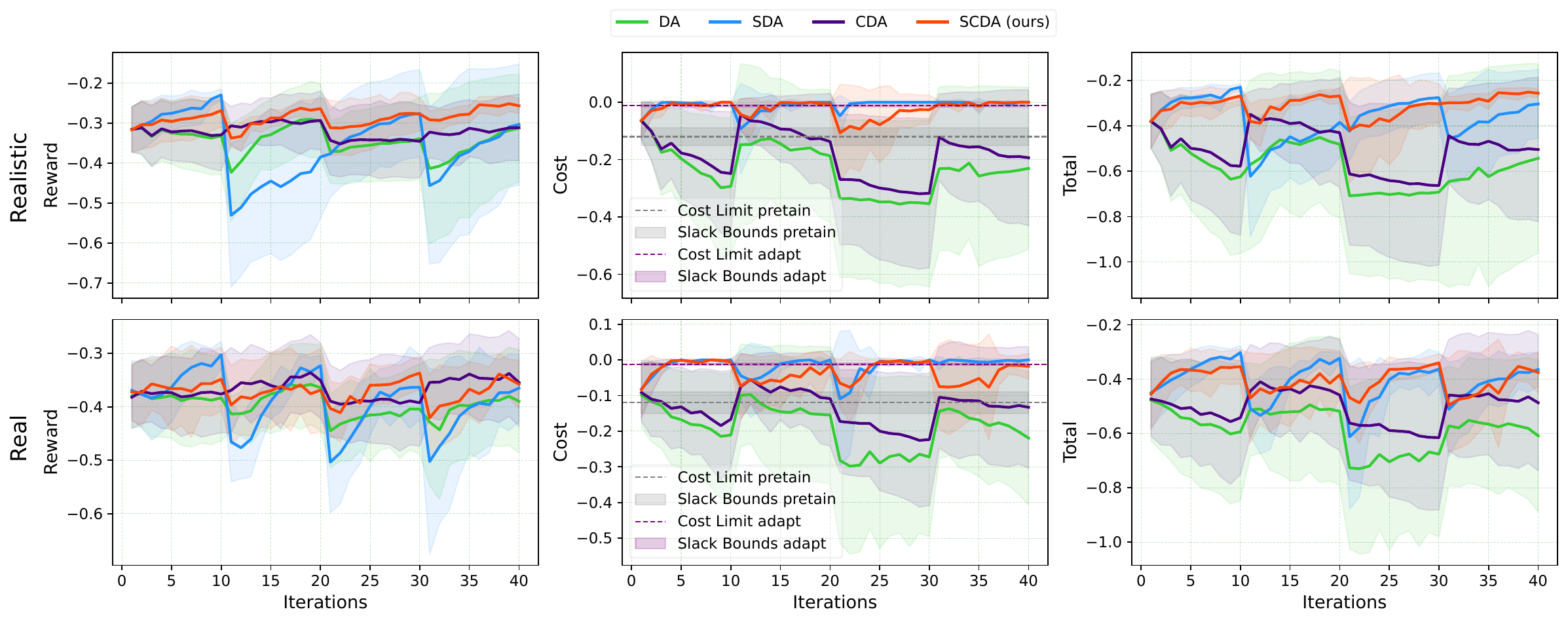}}
    \caption{Training progress during the pretraining phase (top) and the adaptation phase on the realistic (middle) and real system (bottom) under different adaptation strategies in terms of the average timestep reward (left), average timestep cost (center), and the sum of both (right) while adapting to the current target. The target switches every ten iterations during the adaptation phase. Results are averaged over five seeds.}
    \label{fig:reachr_pretrain_adapt}\end{figure*}

\noindent where $\mathbf{d}_t$ is the Euclidean distance between the target and the end-effector, $\mathbf{T}$ is the episode duration, and $\mathbf{\textit{S}}$ is the stopping condition that terminates the episode. During the episode, $\mathbf{\textit{S}}$ becomes true if any of the joints reach a predefined safety rotation limit. The goal of the agent is to maximize reward while minimizing safety violations (maximizing negative cost).

\textit{Implementation details}: Figure \ref{fig:reacher_env} presents the simulated and real environments for the task. For the experiments,  we use the VTPRL\footnote{\url{https://github.com/tum-i6/VTPRL}} simulation environment \cite{Josifovski2020Continual}. We use the KUKA LBR iiwa 14 robotic manipulator \cite{Kuka}. The simulated manipulator meshes are obtained from the URDF data provided by ROS-Industrial \cite{ROSindustrial}. The real robot is controlled with the IIWA stack \cite{Hennersperger2017MRIBased} via \gls{ROS} \cite{ros-operating-system}. One timestep is $100ms$, and an episode lasts up to 10 seconds (100 timesteps) if no safety violation occurs. The maximal joint velocity is restricted to 20 $deg/sec$, and the allowed range is $\pm150deg$ for joint1 and $\pm80deg$ for joint2. The robot starts from an upright position where both joints are at 0 degrees. For domain randomization purposes, similarly to \cite{Peng2018SimtoReal, josifovski2022analysis, josifovski_auddy2024continual}, we randomize the torques of the robot joints, and we incorporate random noise in the observations of the agent. The policy, the reward, and the cost Critic networks are MLPs with two hidden layers of 64 units each. The value of $\lambda$ is set to 1 during the experiments for strategies that use continual learning and 0 otherwise. The default cost limit and slack bound are set to 0.12 and 0.03, respectively, while all other PCRPO parameters are used as defined by Gu et al.\ \cite{gu2024balance} for the 4S-G-V0 variant. All hyperparameters are also available at \url{https://safe-cda.github.io/}.  

\textit{Pretraining Phase}: During this phase, the agent learns from scratch how to reach and balance over 4 targets in a highly parallelized simulation with 64 episodes per batch, where each batch update contains samples from the different targets, representing the general domain distribution.
We use wide randomization ranges in the pretraining phase in order to train a robust model (20\% noise on the observation, stiffens and damping coefficient values for joint torques are randomized between 10 and 1000). During the optimization, we use the true reward and cost signals from the simulator, as this leads to better training convergence and sim2real transfer in zero-shot settings \cite{malmir2023diarel}. We perform 200 batch updates until the training converges. Finally, we take a snapshot of the policy parameters and calculate their Fisher importance as explained in Section \ref{sec:method}.

\textit{Adaptation Phase}: For the adaptation phase
we consider two different transfer domains, either the real system or a realistically simulated system based on the real system \footnote{For the realistic system, we set the observation noise between 0 to 2 percent based on the manufacturer's manual on torque reproducibility. The stiffness and damping coefficients were set to 10 and 20, respectively, by comparing the simulated and the real torque profiles. Both systems showed similar trends, but we keep the realistic system and report both results to facilitate reproducibility.}.
During the adaptation phase, the agent is adapted sequentially on only one target at a time. Here, we use a stricter safety limit to 10\% of the one in the pertaining phase to minimize unsafe exploratory actions. We sample 10 episodes per batch and perform 10 batch updates of the model per target. In this case, we optimize for the (possibly) noisy or delayed reward/cost, as would be the case in the real world, where the agent cannot access a perfect reward signal. After adapting the policy to each of the targets, we take snapshots and test how well it is performing on the target it was adapted to vs.\ the ones it was not actively adapting to. This sequential adaptation process allows us to test the dynamics shift from the source domain (randomized simulation) to the target system (realistic or real system) and how the subdomain distribution affects the performance under different transfer strategies.

We consider the following transfer strategies:

\begin{itemize}
	\item \textit{Zero-shot Transfer}\textit{(ZS)} -- The pre-trained model in randomized simulation is directly deployed to the real(istic) system and used in a zero-shot manner. Serves as a baseline. 
 
	\item \textit{Domain Adaptation}\textit{(DA)} --  A model pre-trained in randomized simulation is further adapted on the real(istic) system sequentially on each of the targets while only optimizing for the reward. 

	\item \textit{Safe Domain Adaptation}\textit{(SDA)} -- Similar to DA, but during the adaptation phase optimization is done for both safety and reward.

	\item \textit{Continual Domain Adaptation}\textit{(CDA)} --  Similar to DA, but during the adaptation phase, policy optimization is regularized to stay close to the initial pre-trained policy.   
	
	\item \textit{Safe Continual Domain Adaptation}\textit{(SCDA)} -- Similar to CDA, but the policy is optimized both on safety and reward criteria while staying close to the original pre-trained policy. 
\end{itemize}

\subsection{Experiment Results}

\begin{figure*}[t]
    \centering
    \includegraphics[width=\linewidth]{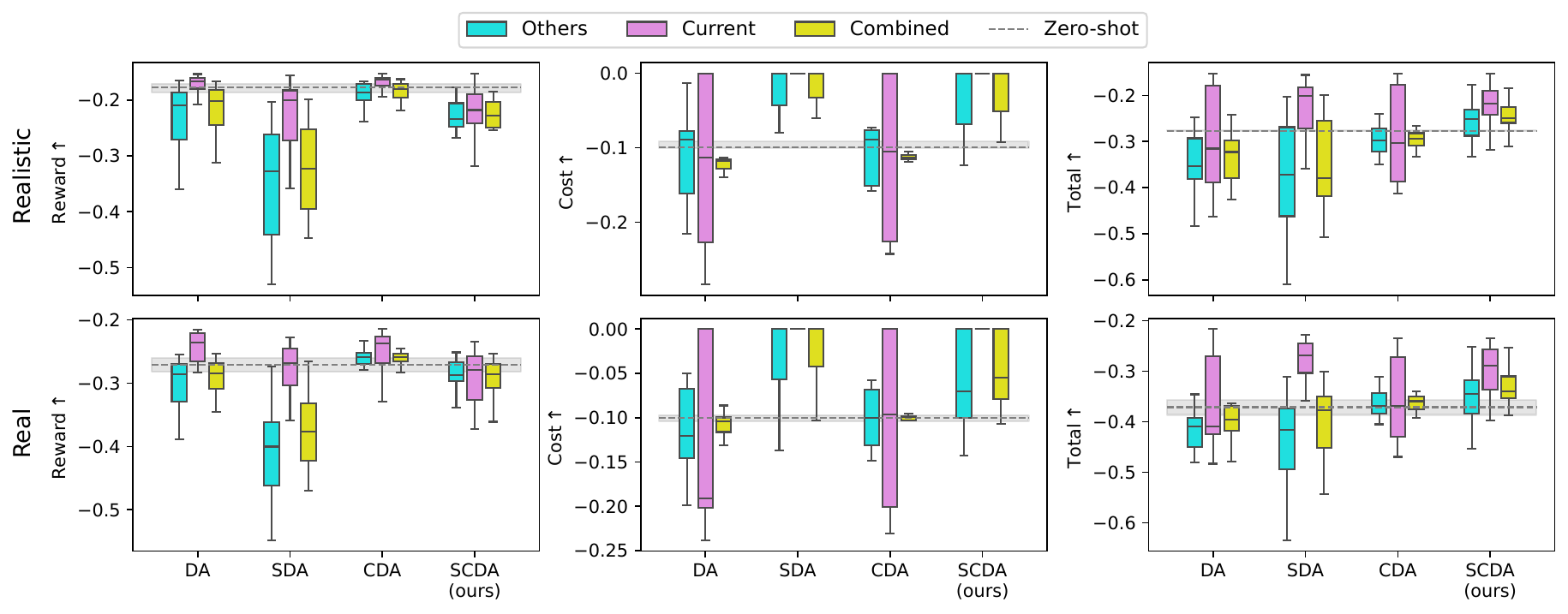}
    \caption{Realistic (top) and real system (bottom): Performance of different adaptation strategies in terms of the episode reward (left), cost (center), and total (right) while adapting to the current target (Current) compared to performance on the other targets (Others). Results are shown for five seeds.
    }
    \label{fig:inference_results_real}
    \vspace{-15px}
\end{figure*}

The results of the pretraining phase and the adaptation phase are presented in Figure \ref{fig:reachr_pretrain_adapt}. In the pertaining phase, the agent learns to perform the task in randomized simulation while adhering to the predefined average cost limits per timestep (top row). In the adaptation phase, the strategies that do not optimize safety (DA and CDA) lead to a slight improvement in the reward (mean reward value at switching to a specific target vs.\ mean value after 10 iterations on it in the reward subplots), however this results in overshooting the target and activating the safety limits, and they are not able to recover from this (cost subplots). The strategies that use safety optimization (SDA and SCDA) can, on average, stay within the predefined safety limits of the pretraining phase and optimize the policy to the much stricter 10\% safety bounds defined under the adaptation phase, while also improving the reward once the safety bounds are not violated. The SDA strategy, under which the policy is not restricted to staying close to the initial pre-trained policy, can adapt faster when a target switch occurs, unlike SCDA, where the change is more gradual. However, we can also notice an overfitting trend with SDA under the current distribution (evident in the reward plot for the real system). At the target-switching time, it has to start optimizing the reward for the current target from a much worse value than the other strategies. 

While the above results give us a clearer picture of the ability of different strategies to allow safe exploration and adaptation to the transferred domain's current dynamics and distribution in training mode, in Figure \ref{fig:inference_results_real}, we see the results for different strategies in inference mode, i.e., when the agent is taking the optimal actions under the current policy. We consider three different cases:
\begin{inparaenum}[(i)]
    \item \textit{Current}: Average policy performance on the target it was actively adapted to. For example, what is the performance of the policy snapshot at iteration 10 on Target 1 or the policy snapshot at iteration 20 on Target 2? 
    \item \textit{Others}: Average policy performance on the targets it was not actively adapted to. For example, what is the average performance of the policy snapshot at iteration 10 on Targets 2, 3, and 4? 
    \item \textit{Combined}: The average performance of the policy on all the targets after being actively adapted on only one, i.e., after 10, 20, 30, and 40 adaptation iterations.  
\end{inparaenum}

From analyzing the cost plots, we see that the pre-trained policy (ZS transfer represented by the shaded area) does not transfer perfectly to the transfer domain -- for some of the targets, it generates cost by reaching joint safety limits. When continuing to optimizing for safety (SDA and SCDA), we can minimize the cost under the current distribution and improve the overall cost w.r.t.\ the pre-trained policy. In practice, this means that the agent commands slower speeds to the robot when approaching the target or avoids approaching it, since it is close to a safety violation value for the second joint. This comes at the price of reducing the gain in reward since the reward is inversely proportional to the distance from the target. On the other hand, if we consider the reward subplots, we see that by only optimizing for the reward (DA and CDA), we can improve the reward under the current distribution, but this leads to very unsafe actions, as the agent learns to speed up and overshoot the target, leading to safety limits violation.

By analyzing the \textit{total} plot, which shows the overall performance of the strategies when we combine cost and reward, we see that both SDA and SCDA can improve over the pre-trained policy (ZS transfer) when adapting to the current domain distribution. However, the SDA strategy, which does not use continual learning, overfits to the current domain distribution and dynamics, as is evident from the discrepancy in its performance on current vs.\ other targets. In this case, SDA would take more adaptation steps when the current domain dynamics and distribution shifts. This could be especially problematic under periodic domain shifts, where the system would have to switch to training mode regularly and undergo longer adaptation phases. On the other hand, SCDA improves the performance under the current distribution while not forgetting the general pre-trained policy, as is evident by the slight difference in total performance between its current and other targets. In addition, if we look at the total performance of SCDA over all the targets, we can answer Q1 positively, as it is the only strategy that can improve the total performance over the whole domain w.r.t.\ the zero-shot general policy. 

\begin{figure}
    \centering
    \includegraphics[width=0.48\textwidth]{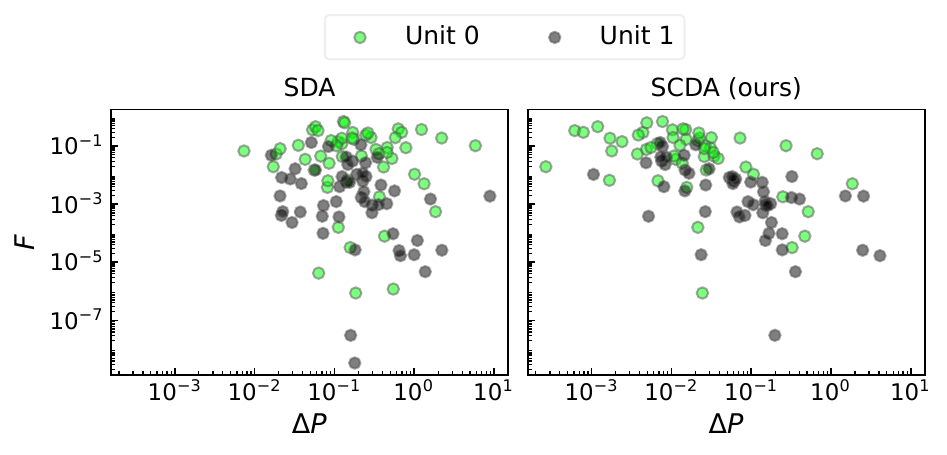}
    \vspace{-20px}
    \caption{Relative parameter change as a function of Fisher importance.}
    \label{fig:fisher}
    \vspace{-20px}
\end{figure}

Further analysis of the difference in adaptivity of the policy under SCDA and SDA can be done by comparing the changes of the policy parameter values after adaptation w.r.t.\ their values at the end of the pretraining phase. As an example, the plots in Figure \ref{fig:fisher} show the relative change (x-axis) of each of the 128 weights that connect the last hidden layer with the two units of the output layer as a function of their Fisher importance in randomized simulation (y-axis). We can see that under the SDA strategy, the changes are more significant and uncorrelated to their Fisher importance, while under SCDA, the higher their Fisher importance, the lower the relative change is. This enables SCDA to ``remember'' the general policy found in randomized simulation while changing the less important policy parameters to fit the current subdomain and dynamics. In addition, it prevents the policy from diverging too far from the pre-trained one, thus implicitly regulating the exploratory actions in the transferred domain. Nevertheless, this can also reduce the policy's adaptability, as shown in Figure \ref{fig:reachr_pretrain_adapt}, where it takes longer to adapt to a safe policy under SCDA in cases where the pre-trained policy is underperforming. In this case, the $\lambda$ parameter can be used to control the trade-off between adaptability and remembering.

\subsection{Verification on a complex task}
While the reach-and-balance task allowed detailed analysis in a simplified scenario without the risk of collisions when performing experiments on the real system, we applied the SCDA framework to a more complex task involving grasping. Figure \ref{fig:method} presents the simulation and the real-system setup for the task. The task is extended from \cite{petropoulakis2024state}, and the agent should learn to guide the end-effector in a closed-loop control to an optimal grasping position for an object while avoiding joint limits and collisions with the stand. The agent state at each timestep is:
\begin{equation}
s_t = [{\Delta}x_t, {\Delta}y_t, {\Delta}z_t, {\Delta}rz_t, q_{1_t}, ..., q_{7_t}] \in \mathbb{R}^{11} ,
\end{equation}
where ${\Delta}x_t$ and ${\Delta}y_t$ and ${\Delta}z_t$ are the positional errors and ${\Delta}rz_t$ is the rotational error between the end-effector and the object along the corresponding axes, and $q_{i_t}$ is the i-th joint position of the manipulator. The agent has 4DoF control of the actions $a_t$ at each timestep, where $a_t$:
\begin{equation}
a_t = [v_{x_{t}}, v_{y_{t}}, v_{z_{t}}, v_{zr_{t}}] \in \mathbb{R}^{4}
\end{equation}
represents the requested velocities of the end-effector along each axis for position and the z-axis for rotation. The reward and cost terms are defined as:
\small
\begin{align}
\mathbf{r}_t &=
     \begin{cases}
  \Delta{d}_{x_{t}} + \Delta{d}_{y_{t}} + \Delta{d}_{z_{t}} + \Delta{d}_{rz_{t}}
  \\
   0, 
  \end{cases}
\mathbf{c}_t =
    \begin{cases}
   0 & \text{if } \neg S  \\
   C_s, & \text{if } S 
  \end{cases} ,
    \in \mathbb{R}
\end{align}
\normalsize
where $\Delta{d}_{x_{t}}$, $\Delta{d}_{y_{t}}$, $\Delta{d}_{z_{t}}$ and $\Delta{d}_{rz_{t}}$ are the normalized differences of the positional and rotational error from the end-effector to the object along the axes at the current timestep compared to the previous one. Thus, the reward gets higher when moving in the right direction \cite{petropoulakis2024state}.

\textit{Implementation details}: We use the same hardware and simulation environment as before, in addition to the 3-Finger Robotic Gripper by Robotiq \cite{gripper}. The object is a box with a side of $5cm$, placed on a stand at a height of $20cm$. Joint limits are set to $10deg$ before the robot's mechanical limits for each joint. The timestep duration is $50ms$ and each episode is 400 timesteps (20 seconds) long. In case of collision, the stopping condition $S$ is activated -- the robot stops, and a one-time collision penalty ($C_s$) of $-100$ is given. If no collision happens, the gripper is closed and lifted at the end of the episode to check for grasp success. During the pertaining phase, the object is spawned at a random x and y position in the reachable area in front of the robot. For domain randomization, uniform noise in the range of [0, 5\%] is added to the observed state at each timestep. We train for 200 batch updates, each containing 10 episodes. Similarly to the previous task, we calculate the Fisher importance for each policy parameter at the end of the pretraining phase and continue adapting the policy on the real(istic) system. During the adaptation phase, we select one position of the object and do 10 batch updates, each comprising 10 episodes. All other hyperparameters and network architectures are as for the reaching task.

\begin{table}[t]
    \centering
    \caption{Grasping task results over five seeds.}
    \label{table:experimentResultsGrasper}
    \begin{tabular}{@{}l@{\hspace{2mm}}c@{\hspace{2mm}}c@{\hspace{2mm}}c@{\hspace{2mm}}c@{\hspace{2mm}}c@{\hspace{2mm}}c@{}}
\toprule
 & \multicolumn{2}{c}{\hspace{-2mm}Avg.~$r\uparrow$} & \multicolumn{2}{c}{\hspace{-2mm}Avg.~cost $\uparrow$} & \multicolumn{2}{c@{}}{\hspace{-2mm}Grasp succ.~$\uparrow$} \\
Strategy & Realistic & Real & Realistic & Real & Realistic & Real \\
\midrule
Zero-Shot & 0.155$\pm$0.011 & 0.126$\pm$0.055 & -0.1 & -0.1 & 20\% & 20\% \\
SCDA  & \textbf{0.163$\pm$0.021} & \textbf{0.154$\pm$0.022} & \textbf{0.0} & \textbf{0.0} & \textbf{40\%} & \textbf{60\%} \\
\bottomrule
\end{tabular}

\end{table}

After the adaptation, we compare the pre-trained and the SCDA-adapted policy in inference mode. The results of the average timestep reward and cost, as well as the average episode grasp success rate, are presented in Table \ref{table:experimentResultsGrasper} and show that SCDA can significantly improve over zero-shot transfer, leading to higher reward and success rate, while minimizing safety violations. Detailed demonstrations of this task are available in the supplementary video material. 

\section{Conclusion and Future Work}

We presented SCDA, an approach to enable safe continual adaptation of the RL control policy after deployment on a robotic system to compensate for environment and system changes over time. While SCDA offers the possibility of adaptation after deployment, there are some assumptions/limitations that we want to address in future work. One assumption is that the domain shifts on the real system are captured within the randomization ranges of the pre-trained policy. However, it is currently not possible to detect domain shifts or out-of-distribution scenarios automatically. We plan to extend SCDA with switching mechanisms to handle these conditions automatically in future work. Additionally, during initial training, the learning process may violate safety constraints. However, in future work, we aim to ensure safety even in the early stages of policy learning by incorporating control and formal methods, thereby enabling applications with higher safety requirements.

\bibliographystyle{IEEEtran}
\bibliography{bib_merged_cleaned}

\begin{thebibliography}{10}
\providecommand{\url}[1]{#1}
\csname url@rmstyle\endcsname
\providecommand{\newblock}{\relax}
\providecommand{\bibinfo}[2]{#2}
\providecommand\BIBentrySTDinterwordspacing{\spaceskip=0pt\relax}
\providecommand\BIBentryALTinterwordstretchfactor{4}
\providecommand\BIBentryALTinterwordspacing{\spaceskip=\fontdimen2\font plus
\BIBentryALTinterwordstretchfactor\fontdimen3\font minus
  \fontdimen4\font\relax}
\providecommand\BIBforeignlanguage[2]{{%
\expandafter\ifx\csname l@#1\endcsname\relax
\typeout{** WARNING: IEEEtran.bst: No hyphenation pattern has been}%
\typeout{** loaded for the language `#1'. Using the pattern for}%
\typeout{** the default language instead.}%
\else
\language=\csname l@#1\endcsname
\fi
#2}}

\bibitem{muratore2022robot}
F.~Muratore, F.~Ramos, G.~Turk, W.~Yu, M.~Gienger, and J.~Peters, ``Robot
  {{Learning}} from {{Randomized Simulations}}: {{A Review}},'' \emph{Frontiers
  in Robotics and AI}, vol.~9, no. 799893, 2022.

\bibitem{josifovski2022analysis}
J.~Josifovski, M.~Malmir, N.~Klarmann, B.~L. Zagar, N.~{Navarro-Guerrero}, and
  A.~Knoll, ``Analysis of {{Randomization Effects}} on {{Sim2Real Transfer}} in
  {{Reinforcement Learning}} for {{Robotic Manipulation Tasks}},'' in
  \emph{{{IEEE}}/{{RSJ International Conference}} on {{Intelligent Robots}} and
  {{Systems}} ({{IROS}})}, Kyoto, Japan, 2022, pp. 10\,193--10\,200.

\bibitem{muratore2021data}
F.~Muratore, C.~Eilers, M.~Gienger, and J.~Peters, ``Data-{{Efficient Domain
  Randomization}} with {{Bayesian Optimization}},'' \emph{IEEE Robotics and
  Automation Letters}, vol.~6, no.~2, pp. 911--918, 2021.

\bibitem{jakobi1995noise}
N.~Jakobi, P.~Husbands, and I.~Harvey, ``Noise and the {{Reality Gap}}: {{The
  Use}} of {{Simulation}} in {{Evolutionary Robotics}},'' in \emph{European
  {{Conference}} on {{Artificial Life}} ({{ECAL}})}, ser. {{LNCS}}, vol. 929,
  1995, pp. 704--720.

\bibitem{Tobin2017Domain}
J.~Tobin, R.~Fong, A.~Ray, J.~Schneider, W.~Zaremba, and P.~Abbeel, ``Domain
  {{Randomization}} for {{Transferring Deep Neural Networks}} from
  {{Simulation}} to the {{Real World}},'' in \emph{{{IEEE}}/{{RSJ International
  Conference}} on {{Intelligent Robots}} and {{Systems}} ({{IROS}})}, 2017, pp.
  23--30.

\bibitem{josifovski2018object}
J.~Josifovski, M.~Kerzel, C.~Pregizer, L.~Posniak, and S.~Wermter, ``Object
  {{Detection}} and {{Pose Estimation Based}} on {{Convolutional Neural
  Networks Trained}} with {{Synthetic Data}},'' in \emph{{{IEEE}}/{{RSJ
  International Conference}} on {{Intelligent Robots}} and {{Systems}}
  ({{IROS}})}, Madrid, Spain, 2018, pp. 6269--6276.

\bibitem{Zhao2020SimtoReal}
W.~Zhao, J.~P. Queralta, and T.~Westerlund, ``Sim-to-{{Real Transfer}} in
  {{Deep Reinforcement Learning}} for {{Robotics}}: {{A Survey}},'' in
  \emph{{{IEEE Symposium Series}} on {{Computational Intelligence}}
  ({{SSCI}})}, Canberra, ACT, Australia, 2020, pp. 737--744.

\bibitem{kaspar2020sim2real}
M.~Kaspar, J.~D. Mu{\~n}oz~Osorio, and J.~Bock, ``{{Sim2Real Transfer}} for
  {{Reinforcement Learning Without Dynamics Randomization}},'' in
  \emph{{{IEEE}}/{{RSJ International Conference}} on {{Intelligent Robots}} and
  {{Systems}} ({{IROS}})}, Las Vegas, NV, USA, 2020, pp. 4383--4388.

\bibitem{Mehta2020active}
B.~Mehta, M.~Diaz, F.~Golemo, C.~J. Pal, and L.~Paull, ``Active {{Domain
  Randomization}},'' in \emph{Conference on {{Robot Learning}} ({{CoRL}})},
  vol. 100.\hskip 1em plus 0.5em minus 0.4em\relax Osaka, Japan: PMLR, 2020,
  pp. 1162--1176.

\bibitem{OpenAI2019Solving}
OpenAI, I.~Akkaya, M.~Andrychowicz, M.~Chociej, M.~Litwin, B.~McGrew,
  A.~Petron, A.~Paino, M.~Plappert, G.~Powell, R.~Ribas, J.~Schneider,
  N.~Tezak, J.~Tworek, P.~Welinder, L.~Weng, Q.~Yuan, W.~Zaremba, and L.~Zhang,
  ``Solving {{Rubik}}'s {{Cube}} with a {{Robot Hand}}, Tech. Rep.
  arXiv:1910.07113, Oct. 2019.

\bibitem{tiboni2023dropo}
G.~Tiboni, K.~Arndt, and V.~Kyrki, ``{{DROPO}}: {{Sim-to-Real Transfer}} with
  {{Offline Domain Randomization}},'' \emph{Robotics and Autonomous Systems},
  vol. 166, p. 104432, 2023.

\bibitem{gu2024safe}
S.~Gu, D.~Huang, M.~Wen, G.~Chen, and A.~Knoll, ``Safe {{Multiagent Learning}}
  with {{Soft Constrained Policy Optimization}} in {{Real Robot Control}},''
  \emph{IEEE Transactions on Industrial Informatics}, vol.~20, no.~9, pp.
  10\,706--10\,716, 2024.

\bibitem{gu2024review}
S.~Gu, L.~Yang, Y.~Du, G.~Chen, F.~Walter, J.~Wang, and A.~Knoll, ``A
  {{Review}} of {{Safe Reinforcement Learning}}: {{Methods}}, {{Theories}}, and
  {{Applications}},'' \emph{IEEE Transactions on Pattern Analysis and Machine
  Intelligence}, vol.~46, no.~12, pp. 11\,216--11\,235, 2024.

\bibitem{achiam2017constrained}
J.~Achiam, D.~Held, A.~Tamar, and P.~Abbeel, ``Constrained {{Policy
  Optimization}},'' in \emph{International {{Conference}} on {{Machine
  Learning}} ({{ICML}})}, vol.~70, Sydney, NSW, Australia, 2017, pp. 22--31.

\bibitem{tessler2018reward}
C.~Tessler, D.~J. Mankowitz, and S.~Mannor, ``Reward {{Constrained Policy
  Optimization}},'' in \emph{International {{Conference}} on {{Learning
  Representations}} ({{ICLR}})}, New Orleans, LA, USA, 2019.

\bibitem{gu2023safe}
S.~Gu, J.~Grudzien~Kuba, Y.~Chen, Y.~Du, L.~Yang, A.~Knoll, and Y.~Yang, ``Safe
  {{Multi-Agent Reinforcement Learning}} for {{Multi-Robot Control}},''
  \emph{Artificial Intelligence}, vol. 319, p. 103905, 2023.

\bibitem{valassakis2020crossing}
E.~Valassakis, Z.~Ding, and E.~Johns, ``Crossing the {{Gap}}: {{A Deep Dive}}
  into {{Zero-Shot Sim-to-Real Transfer}} for {{Dynamics}},'' in
  \emph{{{IEEE}}/{{RSJ International Conference}} on {{Intelligent Robots}} and
  {{Systems}} ({{IROS}})}, Las Vegas, NV, USA, 2020, pp. 5372--5379.

\bibitem{huang2023what}
P.~Huang, X.~Zhang, Z.~Cao, S.~Liu, M.~Xu, W.~Ding, J.~Francis, B.~Chen, and
  D.~Zhao, ``What {{Went Wrong}}? {{Closing}} the {{Sim-to-Real Gap Via
  Differentiable Causal Discovery}},'' in \emph{Conference on {{Robot
  Learning}} ({{CoRL}})}, Atlanta, GA, USA, 2023.

\bibitem{muratore2022neural}
F.~Muratore, T.~Gruner, F.~Wiese, B.~Belousov, M.~Gienger, and J.~Peters,
  ``Neural {{Posterior Domain Randomization}},'' in \emph{Conference on {{Robot
  Learning}} ({{CoRL}})}, vol. 164, London, UK, 2021, pp. 1532--1542.

\bibitem{parisi2019continual}
G.~I. Parisi, R.~Kemker, J.~L. Part, C.~Kanan, and S.~Wermter, ``Continual
  {{Lifelong Learning}} with {{Neural Networks}}: {{A Review}},'' \emph{Neural
  Networks}, vol. 113, pp. 54--71, 2019.

\bibitem{wang2024comprehensive}
L.~Wang, X.~Zhang, H.~Su, and J.~Zhu, ``A {{Comprehensive Survey}} of
  {{Continual Learning}}: {{Theory}}, {{Method}} and {{Application}},''
  \emph{IEEE Transactions on Pattern Analysis and Machine Intelligence},
  vol.~46, no.~8, pp. 5362--5383, 2024.

\bibitem{aljundi2019task}
R.~Aljundi, K.~Kelchtermans, and T.~Tuytelaars, ``Task-{{Free Continual
  Learning}},'' in \emph{{{IEEE}}/{{CVF Conference}} on {{Computer Vision}} and
  {{Pattern Recognition}} ({{CVPR}})}, Long Beach, CA, USA, 2019, pp.
  11\,246--11\,255.

\bibitem{khetarpal2022towards}
K.~Khetarpal, M.~Riemer, I.~Rish, and D.~Precup, ``Towards continual
  reinforcement learning: A review and perspectives,'' \emph{Journal of
  Artificial Intelligence Research}, vol.~75, pp. 1401--1476, 2022.

\bibitem{schopf_hypernetwork-ppo_2022}
P.~Sch{\"o}pf, S.~Auddy, J.~Hollenstein, and A.~{Rodr{\'i}guez-S{\'a}nchez},
  ``Hypernetwork-{{PPO}} for {{Continual Reinforcement Learning}},'' in
  \emph{Deep {{Reinforcement Learning Workshop NeurIPS}} 2022}, Virtual Event,
  2022.

\bibitem{abel2023definition}
D.~Abel, A.~Barreto, B.~Van~Roy, D.~Precup, H.~P. {van Hasselt}, and S.~Singh,
  ``A {{Definition}} of {{Continual Reinforcement Learning}},'' in
  \emph{Advances in {{Neural Information Processing Systems}} ({{NIPS}})},
  vol.~36, New Orleans, LA, USA, 2023, pp. 50\,377--50\,407.

\bibitem{lesort2020continual}
T.~Lesort, V.~Lomonaco, A.~Stoian, D.~Maltoni, D.~Filliat, and
  N.~{D{\'i}az-Rodr{\'i}guez}, ``Continual {{Learning}} for {{Robotics}}:
  {{Definition}}, {{Framework}}, {{Learning Strategies}}, {{Opportunities}} and
  {{Challenges}},'' \emph{Information Fusion}, vol.~58, pp. 52--68, 2020.

\bibitem{auddy2023continual}
S.~Auddy, J.~Hollenstein, M.~Saveriano, A.~{Rodr{\'i}guez-S{\'a}nchez}, and
  J.~Piater, ``Continual {{Learning}} from {{Demonstration}} of {{Robotics
  Skills}},'' \emph{Robotics and Autonomous Systems}, vol. 165, p. 104427,
  2023.

\bibitem{auddy2023scalable}
S.~Auddy, J.~Hollenstein, M.~Saveriano, A.~{Rodriguez-Sanchez}, and J.~Piater,
  ``Scalable and {{Efficient Continual Learning}} from {{Demonstration Via}} a
  {{Hypernetwork-Generated Stable Dynamics Model}},'' arXiv, Tech. Rep.
  arXiv:2311.03600, 2024.

\bibitem{guo2024off}
Y.~Guo, Y.~Wang, Y.~Shi, P.~Xu, and A.~Liu, ``Off-{{Dynamics Reinforcement
  Learning}} via {{Domain Adaptation}} and {{Reward Augmented Imitation}},'' in
  \emph{Advances in {{Neural Information Processing Systems}} ({{NIPS}})},
  vol.~37, Vancouver, BC, Canada, 2024, pp. 136\,326--136\,360.

\bibitem{josifovski_auddy2024continual}
J.~Josifovski, S.~Auddy, M.~Malmir, J.~Piater, A.~Knoll, and
  N.~{Navarro-Guerrero}, ``Continual {{Domain Randomization}},'' in
  \emph{{{IEEE}}/{{RSJ International Conference}} on {{Intelligent Robots}} and
  {{Systems}} ({{IROS}})}, Abu Dhabi, United Arab Emirates, 2024, pp.
  4965--4972.

\bibitem{yin2025rapidly}
P.~Yin, T.~Westenbroek, S.~Bagaria, K.~Huang, C.-A. Cheng, A.~Kolobov, and
  A.~Gupta, ``Rapidly {{Adapting Policies}} to the {{Real-World}} via
  {{Simulation-Guided Fine-Tuning}},'' in \emph{{{CoRL Workshop}} on
  {{Mastering Robot Manipulation}} in a {{World}} of {{Abundant Data}}},
  Germany, 2024.

\bibitem{ganie2024online}
I.~Ganie and S.~Jagannathan, ``Online {{Continual Safe Reinforcement
  Learning-Based Optimal Control}} of {{Mobile Robot Formations}},'' in
  \emph{{{IEEE Conference}} on {{Control Technology}} and {{Applications}}
  ({{CCTA}})}, Newcastle upon Tyne, United Kingdom, 2024, pp. 519--524.

\bibitem{chen2021context}
B.~Chen, Z.~Liu, J.~Zhu, M.~Xu, W.~Ding, L.~Li, and D.~Zhao, ``Context-{{Aware
  Safe Reinforcement Learning}} for {{Non-Stationary Environments}},'' in
  \emph{{{IEEE International Conference}} on {{Robotics}} and {{Automation}}
  ({{ICRA}})}, Xi'an, China, 2021, pp. 10\,689--10\,695.

\bibitem{cao2024simplex}
H.~Cao, Y.~Mao, Y.~Cai, L.~Sha, and M.~Caccamo, ``Simplex-enabled {{Safe
  Continual Learning Machine}}, Tech. Rep. arXiv:2409.05898, Oct. 2024.

\bibitem{gu2024balance}
S.~Gu, B.~Sel, Y.~Ding, L.~Wang, Q.~Lin, M.~Jin, and A.~Knoll, ``Balance
  {{Reward}} and {{Safety Optimization}} for {{Safe Reinforcement Learning}}:
  {{A Perspective}} of {{Gradient Manipulation}},'' in \emph{{{AAAI
  Conference}} on {{Artificial Intelligence}} ({{AAAI}})}, ser.
  {{AAAI}}'24/{{IAAI}}'24/{{EAAI}}'24, vol.~38.\hskip 1em plus 0.5em minus
  0.4em\relax Vancouver, BC, Canada: AAAI Press, 2024, pp. 21\,099--21\,106.

\bibitem{kirkpatrick_overcoming_2017}
J.~Kirkpatrick, R.~Pascanu, N.~Rabinowitz, J.~Veness, G.~Desjardins, A.~A.
  Rusu, K.~Milan, J.~Quan, T.~Ramalho, A.~{Grabska-Barwinska}, D.~Hassabis,
  C.~Clopath, D.~Kumaran, and R.~Hadsell, ``Overcoming {{Catastrophic
  Forgetting}} in {{Neural Networks}},'' \emph{Proceedings of the National
  Academy of Sciences}, vol. 114, no.~13, pp. 3521--3526, 2017.

\bibitem{kessler2020unclear}
S.~Kessler, J.~{Parker-Holder}, P.~Ball, S.~Zohren, and S.~J. Roberts,
  ``{{UNCLEAR}}: {{A Straightforward Method}} for {{Continual Reinforcement
  Learning}},'' in \emph{{{ICML Workshop}} on {{Continual Learning}}}, vol.
  108.\hskip 1em plus 0.5em minus 0.4em\relax Vienna, Austria: PMLR, 2020.

\bibitem{Josifovski2020Continual}
J.~Josifovski, M.~Malmir, N.~Klarmann, and A.~Knoll, ``Continual {{Learning}}
  on {{Incremental Simulations}} for {{Real-World Robotic Manipulation
  Tasks}},'' in \emph{2nd {{R}}:{{SS Workshop}} on {{Closing}} the {{Reality
  Gap}} in {{Sim2Real Transfer}} for {{Robotics}}}, {Corvallis, OR, USA}, July
  2020, p.~3.

\bibitem{Kuka}
\BIBentryALTinterwordspacing
``Kuka lbr-iiwa.'' [Online]. Available:
  \url{https://www.kuka.com/products/robot-systems/industrial-robots/lbr-iiwa}
\BIBentrySTDinterwordspacing

\bibitem{ROSindustrial}
\BIBentryALTinterwordspacing
``{ROS} {Industrial}.'' [Online]. Available:
  \url{https://github.com/ros-industrial/kuka_experimental}
\BIBentrySTDinterwordspacing

\bibitem{Hennersperger2017MRIBased}
C.~Hennersperger, B.~Fuerst, S.~Virga, O.~Zettinig, B.~Frisch, T.~Neff, and
  N.~Navab, ``Towards {{MRI-Based Autonomous Robotic Us Acquisitions}}: {{A
  First Feasibility Study}},'' \emph{IEEE Transactions on Medical Imaging},
  vol.~36, no.~2, pp. 538--548, 2017.

\bibitem{ros-operating-system}
M.~Quigley, B.~Gerkey, K.~Conley, J.~Faust, T.~Foote, J.~Leibs, E.~Berger,
  R.~Wheeler, and A.~Ng, ``{{ROS}}: {{An Open-Source Robot Operating
  System}},'' in \emph{{{ICRA Workshop}} on {{Open Source Software}}}, vol.~3,
  Kobe, Japan, 2009, p.~6.

\bibitem{Peng2018SimtoReal}
X.~B. Peng, M.~Andrychowicz, W.~Zaremba, and P.~Abbeel, ``Sim-to-{{Real
  Transfer}} of {{Robotic Control}} with {{Dynamics Randomization}},'' in
  \emph{{{IEEE International Conference}} on {{Robotics}} and {{Automation}}
  ({{ICRA}})}, Brisbane, QLD, Australia, 2018, pp. 3803--3810.

\bibitem{malmir2023diarel}
M.~Malmir, J.~Josifovski, N.~Klarmann, and A.~Knoll, ``{{DiAReL}}:
  {{Reinforcement Learning}} with {{Disturbance Awareness}} for {{Robust
  Sim2Real Policy Transfer}} in {{Robot Control}},'' arXiv, Tech. Rep.
  arXiv:2306.09010, 2023.

\bibitem{petropoulakis2024state}
P.~Petropoulakis, L.~Gr{\"a}f, M.~Malmir, J.~Josifovski, and A.~Knoll, ``State
  {{Representations}} as {{Incentives}} for {{Reinforcement Learning Agents}}:
  {{A Sim2Real Analysis}} on {{Robotic Grasping}},'' in \emph{{{IEEE
  International Conference}} on {{Systems}}, {{Man}}, and {{Cybernetics}}
  ({{SMC}})}, Kuching, Malaysia, 2024, pp. 697--704.

\bibitem{gripper}
\BIBentryALTinterwordspacing
``{Robotiq} {Gripper}.'' [Online]. Available:
  \url{http://robotiq.com/products/industrial-robot-hand/}
\BIBentrySTDinterwordspacing

\end{thebibliography}

\end{document}